\documentclass{article}
\usepackage{ijcai20}

\usepackage{times}

\usepackage{soul}
\usepackage{url}
\usepackage[hidelinks]{hyperref}
\usepackage[utf8]{inputenc}
\usepackage[small]{caption}
\usepackage{graphicx}
\usepackage{amsmath}
\usepackage{booktabs}
\usepackage{ifthen}
\usepackage{pdfpages}
\usepackage{csvsimple}
\usepackage{multicol}
\usepackage{subfig}
\usepackage{float}
\graphicspath{{figures/}}

\usepackage{amssymb}
\usepackage{amsmath}
\usepackage{commath}
\DeclareMathOperator*{\argmax}{arg\,max}  

\title{Symmetry as an Organizing Principle for Geometric Intelligence}

\author{
Snejana Shegheva$^1$\footnote{Contact Author}\and
Ashok Goel$^1$\and
\affiliations
$^1$Design \& Intelligence Laboratory, School of Interactive Computing, Georgia Institute of Technology
\emails
sshegheva3@gatech.edu,
ashok.goel@cc.gatech.edu
}

\begin{document}

\maketitle

\begin{abstract}
The exploration of geometrical patterns stimulates imagination and encourages abstract reasoning which is a distinctive feature of human intelligence. In cognitive science, Gestalt principles such as symmetry have often explained significant aspects of human perception. We present a computational technique for building artificial intelligence (AI) agents that use symmetry as the organizing principle for addressing Dehaene's test of geometric intelligence \cite{dehaene2006core}. The performance of our model is on par with extant AI models of problem solving on the Dehaene's test and seems correlated with some elements of human behavior on the same test. 
\end{abstract}

\section{Introduction}

George Polya argued that symmetry plays an important role in the inductive phase of complex problem solving by reducing and ordering the observable facts \cite{polya1990mathematics}. Captivated by visual diagrams of Polya's work in crystallography, M.C. Escher created a \textit{systematic} organization of geometrical transformations and enshrined symmetry as the principal rule underlying his art \cite{escher2004visions}.  Without systematic knowledge of the mathematics governing patterns of symmetry, he created his own "layman's theory" of symmetry, duality, infinity, and paradoxes. Escher comes to \textit{the open gate of mathematics} by exploring how concepts like \textit{repetition}, \textit{rotation} and \textit{reflection} shape our interpretation of boundaries between shapes \cite{haak1976transformation}. 

\begin{figure}[htbp]
\centering
{\includegraphics[width=1.4in,height=1in]{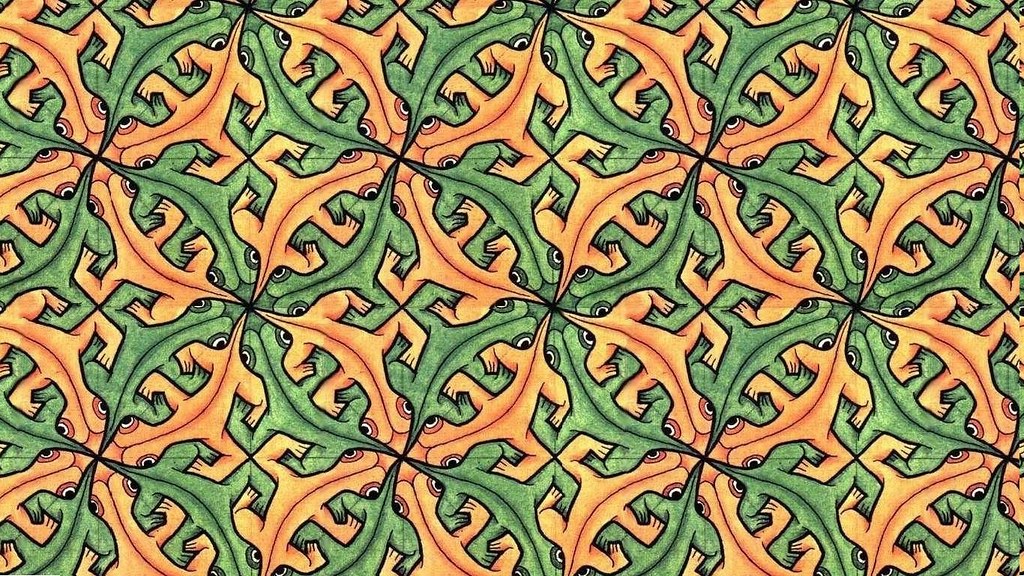}}
\hspace{0.5in}
{\includegraphics[width=1.4in,height=1in]{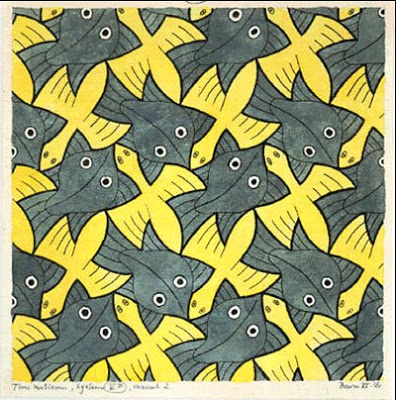}}
\caption{M.C. Escher work inspired by mathematics and nature (Lizards (left) and Bird/Fish (right))}
\label{fig:escher}
\end{figure}

The art work Escher produced over his lifetime profoundly challenges our visual perception of the world. Equally interesting, most humans are capable of understanding and appreciating the beauty of Escher's drawings even in the absence of previous experience with them. Consider, for example, just the two drawings shown in Figure~\ref{fig:escher}. It is easy to see some of the concepts of symmetry -- such as translation, rotation, and reflection -- and this invites questions about the nature of cognitive processes when we perceive this kind of art. 

Indeed, Gestalt psychology has long proposed symmetry as organizing principle of geometric intelligence \cite{bornstein1981perception},\cite{li2009interest}. Gestalt theories suggest that human cognition uses repetition (translational symmetry), foreground/background (rotation and reflection symmetry) in creating and making sense of art \cite{tyler1995empirical}. In sharp contrast, many AI techniques for geometric intelligence rely on object detection as a primary source of power \cite{forbus2011cogsketch}\cite{lovett2008modeling},\cite{santoro2018measuring}. This raises the basic question motivating our work: Might it be possible to build artificial intelligence (AI) agents that use the principles of Gestalt psychology to make inferences about geometric patterns and transformations?  

In an attempt to provide algorithmic answers to these questions, we start with Dehaene's test of geometric intelligence \cite{dehaene2006core} that shares several themes with Escher's more intricate structures. Dehaene \textit{et. al.} describes symmetry as a \textit{geometrical language} that adults and children can comprehend regardless of their culture and background \cite{amalric2017language}. In fact, Dehaene developed the test containing 45 problems to examine whether humans brought up in a technologically advanced civilization with the benefit of formal education including geometry performed better than subjects from a technologically primitive society with little formal education. 

\begin{figure}[htbp]
\centering
{\includegraphics[scale=0.15]{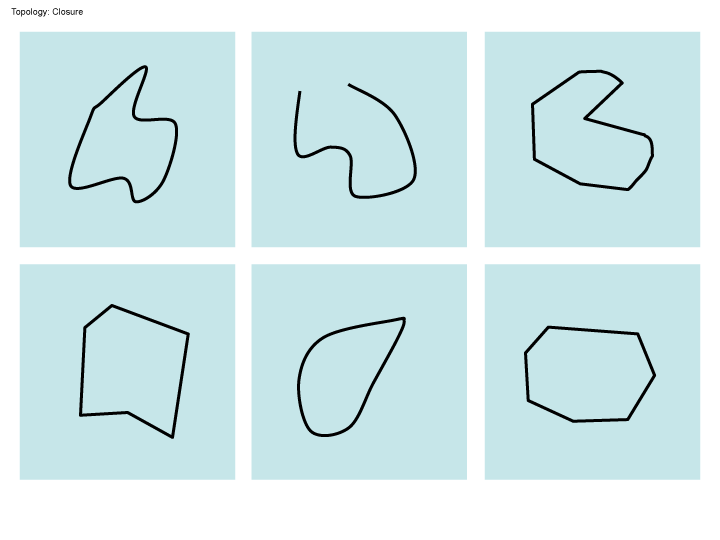}}
\caption{An example of Dehaene's geometry problems that explores topological concepts of closure. One image (here, the top center) violates the concept and therefore should be considered as odd-one-out.}
\label{fig:dehaene}
\end{figure}

 Dehaene's test eschews the use of geometric objects such as triangles and instead relies on more abstract concepts such as \textit{closure}. All 45 problems on the test explore various aspects of core geometry, such as Euclidean geometry, topology, symmetrical figures, metric properties, and geometric transformations \cite{dehaene2006core}. Each problem is an array of six images where one violates the displayed concept, and the test taker is tasked with identifying it as the one that \textit{breaks} the structure.  Figure~\ref{fig:dehaene} shows an example that highlights the above mentioned concept of closure. Although at first glance symmetry is not explicit in Figure 1, we will show below that specific representations of the drawings in Figure 1 derived from Euclidean transformations capture the latent symmetry and order in the drawings. 

Although problems on the Dehaene's test clearly are different from than Escher's more intricate drawings, they nevertheless entail similar, if simpler, kinds of abstract reasoning.  Abstract reasoning on Dehaene's test require the ability to infer higher-level concepts such as relations, symmetries, and complex patterns from low-level pixel representations. According to Dehaene, using geometrical tests with perceptually accessible features such as shapes, positions, and between-object relations, a human capacity to reason abstractly can be measured independent of their culture, language, or experience. This brings us back to the research question motivating our research.  If Gestalt principles underlie human perception, might the same principles form the basis for building AI agents capable of addressing problems on the Dehaene'test? Further, if AI agents based on Gestalt principles indeed can be built for Dehaene's test, what may that tell us about human cognition?  

\section*{Related Work}

Lovett \& Forbus (2017) provide a detailed review of psychological models and studies: A common feature among many of these studies is focus on similarity and especially analogy. Carpenter \textit{et al.} provide a detailed cognitive model of problem solving on the Raven's Progressive Matrices Test of general human intelligence \cite{carpenter1990one}. Their model is based on the traditional production system architecture in which the agent has access to a variety of rules that capture the range of geometric patterns that occur in the Ravens test. 

Lovett, Lockwood \& Forbus (2008) view problems on the Ravens test as geometric analogy problems \cite{lovett2008modeling} and use the structure-mapping theory of analogy to address them. They describe a cognitively inspired approach that detects geometric shapes from an input drawing on the Ravens test, constructs spatial representations of relations among the objects, and then applies the structure-mapping technique for addressing the problem \cite{lovett2008modeling}. 

Kunda, McGreggor \& Goel also view the Raven's test as a set of visual analogy problems. However, in contrast to Carpenter et al. and Lovett et al., they use \textit{affine} transformations, such as translation, rotation and reflection, directly on pixel-level representations to address the Raven's test including the Standard, Color and Advanced Raven's test \cite{kunda2013computational}. Given an input image, their ASTI method interprets the drawing in terms of linear combinations of affine transformations and completes the problem in terms of transformation combination. An interesting aspect of the ASTI computational model is that it does not have prior knowledge of geometric objects and does not need to detect objects, and yet its performance is comparable to that of earlier methods.  

McGreggor, Kunda \& Goel describe a method that makes analogies based on fractal representations. Given an input image, their technique called FAR first builds a fractal representation of the input and then uses similarity and analogy to address the problem. The FAR technique has been successfully used not only for the Ravens test of intelligence, but also for the Odd-One-Out test \cite{mcgreggor2011finding} and the Dehaene's test \cite{mcgreggor2013fractal}. An interesting aspect of the self-similar fractal representations is that the FAR technique can automatically change the level of resolution to best match the given problem. Like ASTI, FAR too does not detect objects in the input images.

In our own earlier work, we have developed a method called Structural Affinity to address problems on the Raven's test of intelligence \cite{shegheva2018structural}. The method uses Markov Random Fields parameterized by affinity factors for first learning the underlying rules as described in Carpenter's work, and subsequently recognizing the pattern in order to make a prediction. The strong emphasis is on discovering topologies that do not rely on object detection, but instead represent features for the type of relationship between images. Not to be confused with image similarity methods, the Structural Affinity captures the \textit{generation} rule that represents the abstract reasoning ability.

As noted above, many previous computational models and techniques have viewed tests of geometric intelligence in terms of extant theories of similarity and analogy. In our current work, we affirm that \textit{symmetry} plays a fundamental role in how an input image is perceived and forms the basis for analogy. Dehaene \textit{et. al.} deliberately minimizes the cues of the concept by randomizing the perceived features. For example, by varying the orientation of the objects or modifying size, the concept becomes obscured. We conjecture that a suitable representation can \textit{undo} the complexity and highlight the right context for the desired concept. 

\section*{Structural Affinity for Core Geometry}
In the current work, we build upon the idea of the \textit{agreement} by constructing different types of affinity factors each representing a property of the geometrical concepts covered in Dehaene's test \cite{shegheva2018structural}. The AI agent \textit{asks} a series of questions from each image and determines the one \textit{weird} image in two steps: 1) identify the most relevant properties that attribute to significant deviations 2) rank images by the largest contribution of variance.  

\subsubsection*{Preliminary Observations}
A distinctive feature of the Dehaene's test is that each problem with six images demonstrates a single concept with small variations. The problem is considered solved if a test taker identifies one out of six images that contains variation not explainable by the concept. Therefore, the aim of the proposed method is to identify the most likely concept by 1) unifying the representation, i.e. reducing superfluous features, and 2) ranking the remaining features by the strength of the signal towards a single pattern. 

We observe that in general, Dehaene's problems exhibit geometrical primitives that fall into specific types of symmetry classes, even in cases where the symmetry concept was not directly targeted.  For example, problems shown in Figures~\ref{fig:dehaene_point} and ~\ref{fig:dehaene_metric}
with intention to highlight the properties of Euclidean geometry, such as distance, can also be interpreted with symmetry - reflective and rotational. Thus, with the assumption of preserving the symmetry, our normalization method rotates and translates the original images re-mapping the pixels to the common axes. This \textit{normalization} intensifies the most relevant features while removing variations intended to obscure the concept. 

We observe that in some instances, the described feature ablation removes information critical to the recognition of the concept. For example, the concept of \textit{chirality} is being erased by the transformation where pixels are reoriented around principal components (see Figure~\ref{fig:dehaene_chiral}). This is a result of the underlying property of chirality - an object cannot be mapped into its mirror image by rotation and translation only. Therefore, there is no symmetry operation (in 2D) that would preserve object's invariance.

\subsubsection*{Image Segmentation Phase}
A Dehaene's problem is represented as a 3x2 matrix of visual entities that capture geometrical shapes and transformations. To represent the problem in the structural affinity framework, we, first, identify the grid and segment the images into its six components. By applying the segmentation, the image is ready to be ingested by our computational model without additional image pre-processing steps. 

\subsubsection*{Representation Phase}
The segmentation phase for one problem produces six panels that are subsequently transformed into an array $A_{i,j}[n \times m]$ where each cell $<i,j>$ takes on a binary value: 1 if the color intensity of the pixel is above a certain threshold $\delta$, and 0, otherwise. Prior to reading the pixels into an array, the image is optionally cropped to remove the pixels associated with a plain text (typically at the top left of the first image).

As the Dehaene's problems target geometry concepts, it makes sense to represent the images in the Cartesian coordinate system instead of a $[n \times m]$  matrix where $n$ and $m$ are the height and the width of the given image. The binary values (0,1) are mapped to real numbers $R^+$. This representation returns a set of points in the coordinate system generated by the Equation~\ref{eq:gx}.

\begin{equation}
f(i,j): Val(A_{i,j}) \mapsto \mathbb{R^+}
\end{equation}

\begin{equation}
g(x) = \{(i,j) \quad | \quad a[i,j] = 1, i \in n, j \in m\}
\label{eq:gx}
\end{equation}

\subsubsection*{Transformation Phase}
\begin{figure}%
    \centering
    \subfloat[Original Figures]{{\includegraphics[width=3cm]{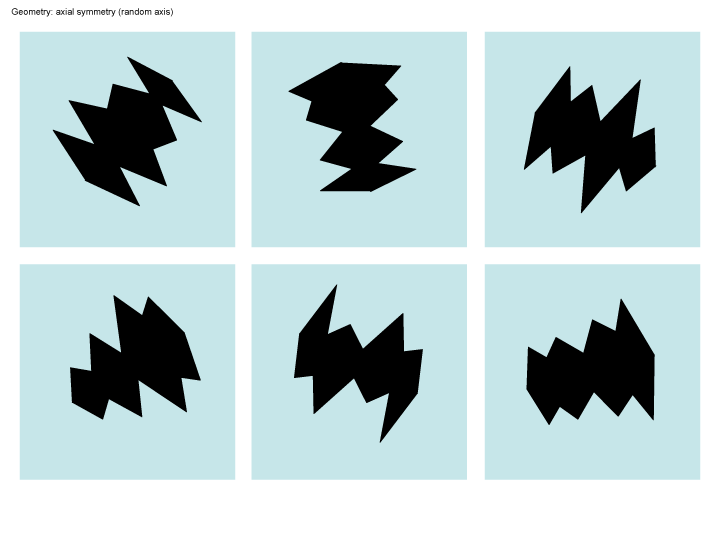}}}%
    \qquad
    \subfloat[Transformed Figures]{{\includegraphics[width=3.5cm]{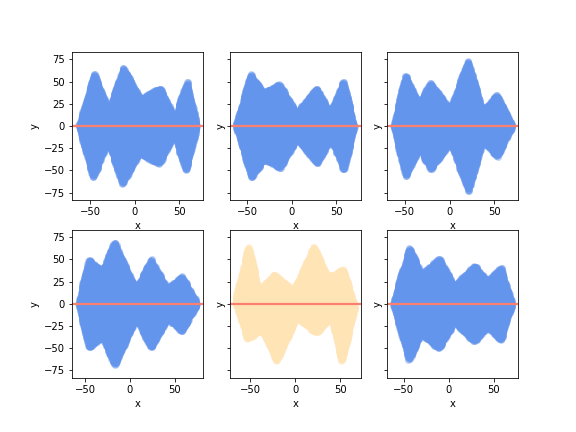} }}%
    \caption{\textbf{Symmetry Concept}. In (a) the concept is shown in its original raw form that include an orientation noise not relevant to the concept. In (b) the transformation applied highlights the single concept of symmetry}%
    \label{fig:dehaene_pca}%
\end{figure}

We use Principal Component Analysis (PCA) method to \textit{unrotate} the figures and obtain the coordinate axes that contain the maximum variance \cite{tipping1999probabilistic}. Figure~\ref{fig:dehaene_pca}(b) shows the result of a successful re-orientation that removes the randomness in the original axes and brings the \textit{symmetry} feature into focus. 

The design of the features is inspired by the classical principles in Gestalt perception - proximity, similarity, common fate, good continuation, closure, symmetry, parallelism \cite{wagemans2012century}. The set of simple heuristics is used for building the knowledge representation. The intentions is to design functions that can serve as cues for the underlying concept. 

\begin{itemize}
    \item \textbf{Color} concept can be encapsulated with a \textit{density} function, that counts number of pixels of varying intensity. During the representation phase, the pixels are transformed into coordinate points, therefore reducing the density function to a count of points.
    
    \item \textbf{Orientation} concept must be captured before the PCA transformation is applied that rotates the figures to the simplest structure. By computing \textit{Pearson} correlation coefficient, we obtain the amount of linear relationships between points \cite{benesty2009pearson}.
    
    \item For \textbf{Topology} concepts, such as \textit{inside/outside}, \textit{closure}, \textit{connectedness}, and \textit{holes}, we design several features - \textit{contour count} and \textit{child-parent} relationship between contours. For example, Figure~\ref{fig:dehaene} that highlights \textit{closure} contains one figure that is inconsistent with the other figures with regards to the number of contours. The \textit{odd} figure shows a disjointed curve whereas the consistent with the concept figures shows curves that join continuous points.  
    
    \item \textbf{Symmetrical Figures} after PCA transformation are automatically aligned along the axes of symmetry, thus computing the discrepancies between the alignment points in the upper and lower quadrants, will give a cue on how symmetrical is the figure. 
    
    First, we \textit{collapse} the points to a single vector by computing the average value per x-coordinate. In symmetrical figures, the vector should contain values close to zero.
    
    \begin{equation}
    \hat{Y}_k = \frac{1}{n_k}\sum_{i=0}^{i=n_k} y_i[x_k] 
    \end{equation}
    
    where $\hat{Y}_k$ is the average value of the points where x=k.
    
    To obtain a scalar measure of the symmetry feature, we subsequently compute the variance of the $\hat{Y}_k$ vector - $Var(\hat{Y}_k)$ that captures the overall adherence to the concept. Likewise, we compute an average vector $\hat{X}_k$ and y-axis along with its scalar representation via variance - $Var(\hat{X}_k)$.
    
    \item The similar features are relevant for \textbf{Geometrical Figures} concept. Certain figures can be mapped into itself in their lines of symmetry. For example, a square has four lines of symmetry, whereas a rectangle has only two. An equilateral triangle has three symmetry lines, whereas an isosceles triangle has only a bilateral symmetry. Designing several functions that capture the heuristics of symmetry nature (reflection, translation, rotation), in conjunction with other features can identify the most inconsistent with the concept figure.     
        
    \item \textbf{Euclidean Geometry} (e.g., line, points, parallelism, and right angle) requires a subset of features defined above. Figure~\ref{fig:dehaene_point} suggests that a feature that computes a variance along axes can easily identify the \textit{weird} image that violates the consistent measure across the remaining figures.
    
    \begin{figure}%
        \centering
        \subfloat[Original Figures]{{\includegraphics[width=3cm]{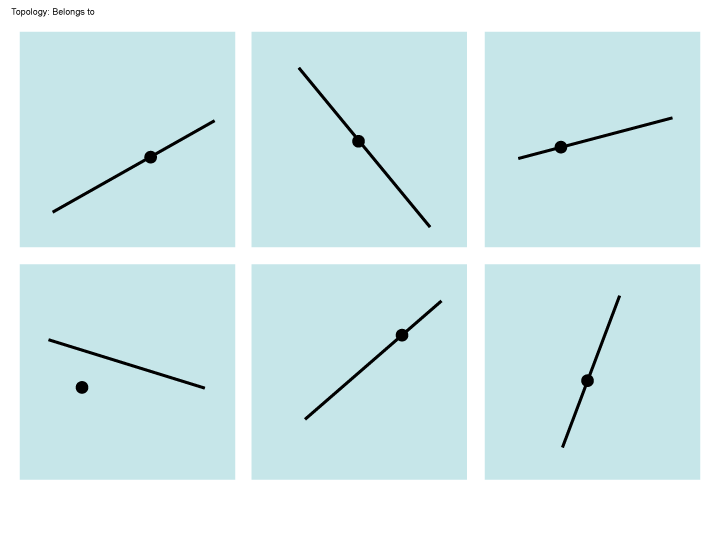}}}%
        \qquad
        \subfloat[Transformed Figures]{{\includegraphics[width=3.5cm]{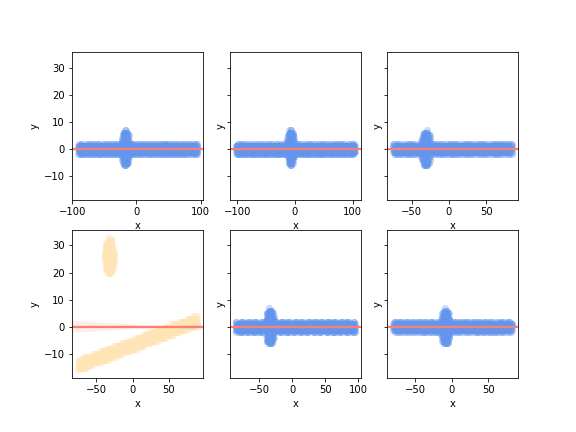} }}%
        \caption{\textbf{Alignment of Points Concept}. In (a) the concept is shown in its original raw form that include an orientation noise not relevant to the concept. In (b) the transformation applied highlights the single concept of alignment}%
        \label{fig:dehaene_point}%
    \end{figure}

    \item Problems that assess the ability to detect \textbf{Metric Properties}, such as distance between objects, center and middle segments, are likewise dependent of the \textit{symmetry} features. In problems where the center of an object is intentionally offset, a feature that creates a mapping between upper and lower quadrant points, exhibits a \textit{gap}, i.e. a number of points from that upper quadrant that do not have a pair from the lower quadrant (see Figure~\ref{fig:dehaene_metric}).
    
    \begin{figure}%
        \centering
        \subfloat[Original Figures]{{\includegraphics[width=3cm]{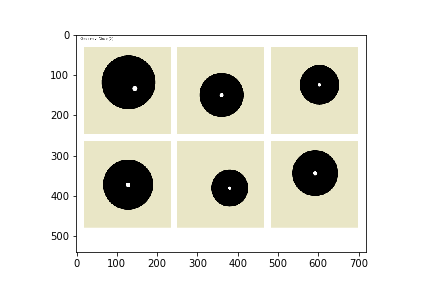}}}%
        \qquad
        \subfloat[Transformed Figures]{{\includegraphics[width=3cm]{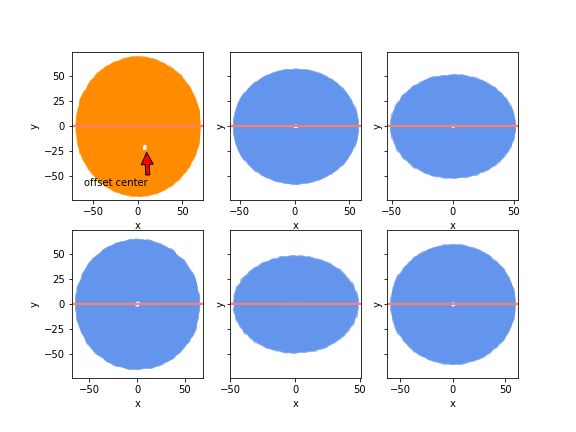} }}%
        \caption{\textbf{Center of Circle}. In (a), the concept is shown in its original raw form that includes circles of different sizes and positions. In (b), the transformation applied highlights the single concept of the location of the center point}%
        \label{fig:dehaene_metric}%
    \end{figure}
    
    \item \textbf{Chiral Figures} require an ability to perform \textit{mental rotation} in order to align the figures in the same axes for comparison. Figure~\ref{fig:dehaene_chiral}(a) shows an example of chiral figures rotated on oblique axes. In this scenario, applying PCA Figure~\ref{fig:dehaene_chiral}(b) removes the information needed to distinguish between the figures making them identical.

    \begin{figure}%
        \centering
        \subfloat[Original Figures]{{\includegraphics[width=3cm]{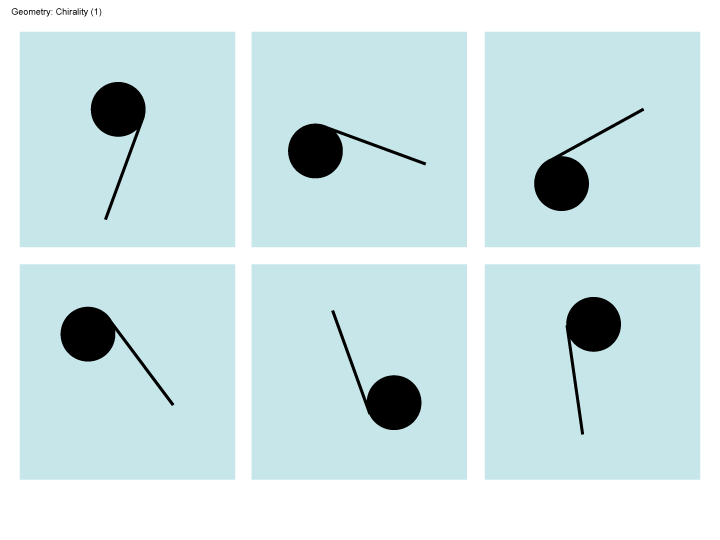}}}%
        \qquad
        \subfloat[Transformed Figures]{{\includegraphics[width=3cm]{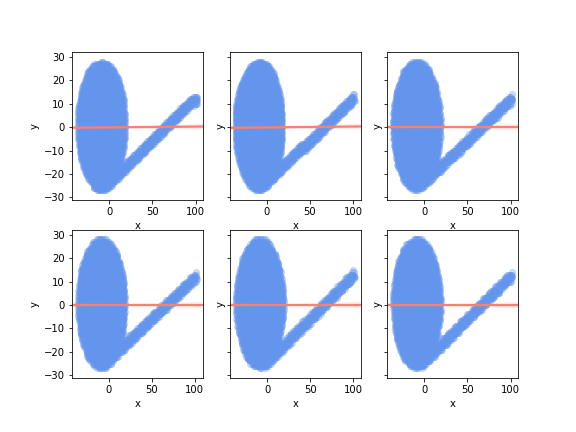} }}%
        \caption{\textbf{Chirality on Oblique Axes}. In (a), the concept is shown in its original raw form that includes chiral figures in random axes. In (b), the transformation applied removed the rotation which made the all the figures identical.}%
        \label{fig:dehaene_chiral}%
    \end{figure}
    
    \item \textbf{Geometrical Transformations} involve \textit{translation}, \textit{homothecy}, \textit{symmetrical reflection}, and  \textit{rotation}. Arguably, this is the most difficult mathematical concept, especially when given in static images \cite{dehaene2006core}. All previously described features are applicable here as well, although the more confident answers are achieved using a combination of several features.
    
\end{itemize}

\subsubsection*{Problem Solving Phase}
\begin{figure*}[htp]
\centering
{\includegraphics[scale=0.5]{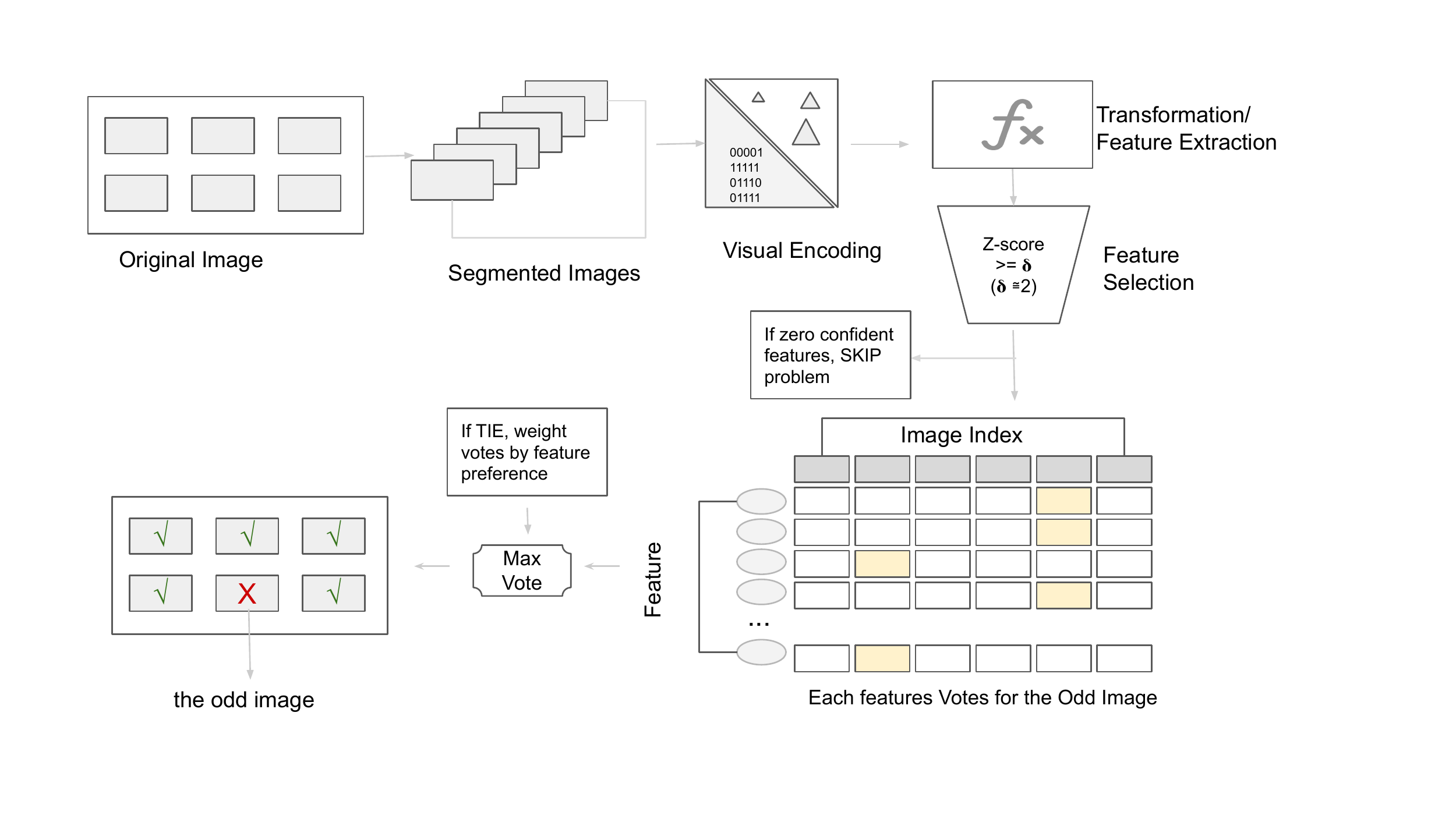}}
\caption{A  diagram  of  the  computational model for solving Dehaene's geometrical problems. The process starts with the original image segmented into six panels which are subsequently encoded and transformed for visual reasoning. Feature selection is performed based on the z-score exceeding a threshold $\delta$ (in our case $\delta = 2$). The discrepant image is identified as the one that violates the observed consistency (voting takes place if more than one feature is selected as a candidate for the underlying concept).}
\label{fig:dehaene_diagram}
\end{figure*}

The general process of our computational model is presented in Figure~\ref{fig:dehaene_diagram}. The algorithm starts with the segmentation and visual encoding as described in the sections above. After the raw features are extracted and transformed, a filtering method is applied to select the most prominent feature(s) $S_i$ that might hold the cues for identifying the discrepant image. 

\begin{equation}
S_i = \{f_k({Im_i^*)} \quad |\quad  \forall f \in Features, \quad z(f_k) \geq \delta \}
\label{eq:features}
\end{equation}

where $f_k(Im_i^*)$ is a $k-th$ feature extracted from all six segmented and encoded images for the original problem $i$;

and $z(f_k) = \frac{x_k - \tilde{x_k}}{\sigma_k}$ - the standard score computed for the feature $z(f_k)$; in our experiments $\delta = 2$ in order to select features where an element is at least two standard deviations from the mean. 

The intuition behind applying a z-score to the raw features is that the core of the Dehaene's test is to detect violations of some desired geometry concept. Therefore, we expect that most pertinent feature will contain an anomaly that can be captured with metrics specifically designed to detect outliers. 
Thus, if the set of figures contain a \textit{weird} attribute, the feature will draw an attention to it with an anomalous z-score. If none of the extracted features exceed the threshold, the extraction process is deemed insufficient, and the problem is skipped.

To identify the discrepant image $X^*$ each of the filtered features \textit{votes} for one of the six candidates, and the solution is selected by choosing the candidate with the highest number of votes. 

\begin{equation}
X^* = \argmax_{j} \quad Vote(X_j)
\label{eq:voting}
\end{equation}

Ideal are scenarios in which filtering yields a single feature, and no voting is required. The ambiguity only arises where more than one feature holds an attribute of the underlying geometry concept. For example, \textit{symmetry} concept can be captured with different high-level heuristic functions, such one-to-one point mapping, or distance to the center line. For \textit{ties}, voting is repeated with a feature preference weight (simpler features are preferred.)

Additionally, the algorithm applies aggressive rounding of the computations across all phases - from representation to transformation, and scoring. This reduction in precision helps with dealing with noise when working with raw images \cite{zadeh1984making}.

\section*{Results}

\subsubsection*{Accuracy}
Table~\ref{tab:concept_result} summarizes the performance of the designed algorithm per metric as presented in the Dehaene et. al. experiment. The total accuracy measure is 89\% with 40 out of 45 problems solved correctly. The last column in the table records the performance of Munduruk\'u participants. Problems involving basic concepts of geometry, such as topology, geometrical figures, lines and angles, have unambiguous features, and therefore are simpler (100\% correct). Problems that involve transformations are more complex as they involve mental translation, reflections, scaling, and rotations. The lower performance on those problems is consistent with the findings reported by Dehaene et. al. Problems that exercise the concept of \textit{chirality} (number of examples = 4) average to a 50\% accuracy with a significant difference between figures shown in vertical vs. random (oblique) axes (~85\% and 23\% correspondingly). 

\begin{table}[ht]
\centering
\begin{tabular}{l|c|c|c|c}
\bfseries Concept & true & total & ratio & human 
\begin{filecontents*}{figures/dehaene_results_per_concept.csv}
Concept,n_correct,total,ratio_correct,munduruku_performance
Color/Orientation,2.0,2.0,1.0,95.0
Chiral Figures Aligned,2.0,2.0,1.0,88.0
Euclidean Geometry,8.0,8.0,1.0,84.0
Geometrical Figures,9.0,9.0,1.0,79.0
Topology,4.0,4.0,1.0,76.0
Symmetrical Figures,3.0,3.0,1.0,67.0
Metric Properties,6.0,7.0,0.86,62.0
Geometrical Transformations,6.0,8.0,0.75,34.0
Chiral Figures Random,0.0,2.0,0.0,23.0
\end{filecontents*}
\csvreader[head to column names]{figures/dehaene_results_per_concept.csv}{}
{\\\hline\csvcoli&\csvcolii&\csvcoliii&\csvcoliv&\csvcolv}
\end{tabular}
\caption{Algorithm performance per concept. The index of the table is the name of the concept as presented by Dehaene et. al.; first column is the count of correctly identified images by the structural affinity algorithm; second column is the total count of problem in the given concept; third column is the ratio of correctly solved problems; And the last column is the performance of the Munduruku participants}
\label{tab:concept_result}
\end{table}

\subsubsection*{Comparison to Other Computational Models}

In this paper, we demonstrate that by leveraging Gestalt principles of perception in the design of algorithms for problem solving, we can approximate the \textit{humanlike} thinking on core geometry tasks. Unlike models that rely on objects and borders detection for qualitative representations \cite{lovett2008modeling}, \cite{lovett2011cultural}, we describe a computational model that 1) \textit{transforms} input images into more perceptually coherent variations, 2) scores the resulting representations against pre-defined properties (such as symmetry, rotation, and other geometrical concepts), and 3) identifies an instance where scores are in disagreement with the rest of images. Lovett at al. \cite{lovett2008modeling} and Lovett and Forbus \cite{lovett2011cultural} integrate four different systems for addressing visual oddity tasks, and solve 39 out of 45 problems correctly. The advantage of our approach is that it solves a similar number of problems correctly (40 out of 45), and it does so while satisfying the \textit{parsimony} characteristic, i.e. striving for the simplest theory that can explain the intent in the discussed geometry problem. Parsimony in this context is concerned with the problem solving behavior, and it serves as a cognitive characteristic of gifted individuals \cite{koichu2008considerations}.

In a computational model that employs \textit{fractal} representation for reasoning \cite{mcgreggor2013fractal}, visual oddity tasks are addressed with a notion of visual similarity that operates on varied levels of image representation - from coarse to refined. In addition to providing solutions, the authors compute confidence measures (30 unambiguously correct, and 15 correct but ambiguous) and compare them with the human performance. An analysis of ambiguity allows tuning the levels of coarseness to determine the best strategy for representation. McGreggor et al. claim that their algorithm is parsimonious because it does not use additional mechanisms for problem solving phases \cite{mcgreggor2013fractal}. However, simplicity is not always a sufficient measure of parsimony; nor does simplicity always offer a robust explanation of the behavior. Our approach is based on scoring images by their explainability with concepts such as symmetry, topology, and Euclidean geometry properties. An image is considered odd if its characteristics do not match characteristic of the rest of the images. An analysis of agreements and violations helps highlight the most pertinent attributes of the problem, thus giving an explainable answer.  

An algorithm of visual perception that is intuitive based on Gestalt principles can infer high level meaning between pixels and their relationships. This in turn increases the capability to \textit{understand} and \textit{reason} in the context of more complex images. 

\section*{Discussion}
One of the main questions raised by Dehaene et. al. is whether or not core geometric intelligence is inherent in the human mind \cite{dehaene2006core}. This question is complex and it depends on how the geometrical concepts are represented, and what type of mental operations are invoked when assessing a concept. Therefore, the research presented here is motivated by two goals: 1) to understand what transformations are ubiquitous for exploring geometrical notions, i.e. what kind of abstract reasoning is most relevant for dissecting a geometry problem, and 2) is there an underlying organizing principle that governs our perception of shapes, and their ability to be transformed into similar, but more perceptually coherent objects. 

To answer (1) we developed a computational model that transforms encoded images that hold geometry concepts using a mathematical mechanism helpful for discovering structure and relations among the variables. We found the applying PCA (principal component analysis) can be interpreted as performing mental transformations, such as rotations of the coordinate axes to align the studied shapes. In the Dehaene's problems, finding the principal components is a viable explanation of how mind removes irrelevant differences between shapes in order to identify the central geometry concept. The resulting structural uniformity brings into focus those aspects that violate the desired geometrical concept. 

To address (2), we observe that most Dehaene's problems exhibit \textit{symmetry} as a latent (unless directly specified) principle. Gestalt theory suggests that visual perception is frequently driven by the tendency to maximize the appeal of the shapes and their connection with other surrounding shapes \cite{spelke1990principles}. Therefore, our computational model benefits from viewing the geometry problems through lens of symmetry. For example, Figure~\ref{fig:dehaene_metric} is probing the concept of \textit{metric properties}, specifically a center of a circle. One way of reasoning about that in concordance with a \textit{metric} concept is to consider the distance of the inside point from all of the circumference points. Alternatively, it may be reasoned with the concept of symmetry - the misalignment of the inner point from the center in one of the figures \textit{breaks} the symmetry group of the circle object, i.e. it is not possible to map the figure on itself using rotations or reflections. Similar reasoning is applied to the other problems in the Dehaene's set suggesting that symmetry may be the underlying organizing principle for basic geometry concepts. 

\section*{Future Directions}
M.C. Escher's  engagement in mathematical ideas through art brings diverse audiences towards a better understanding of the fundamental geometric rules governing complex shapes and transformations \cite{haak1976transformation}. Escher absorbs himself in the playfulness of the \textit{recognizable} figures (motifs) by suppressing the distinction between foreground and background \cite{escher2004visions}. 

The self-organizing processes of the brain described by Gestalt principles enable spontaneous switching between the functions of the figure and ground \cite{wagemans2012century}. Thus,  
it is reasonably easy for a human to identify the motifs in Escher's drawings without requiring substantial previous exposure and training in geometry concepts. In the next stage of our research, we will probe the ability of AI agent to infer the organizational functions in the figure/ground images. 

The main focus of this work was to encode the general properties of grouping principles during perceptions of geometrical primitives in Dehaene's images. In future work, we intend to expand the arsenal of geometrical concepts that can be organized with various types of symmetrical operations and groups. In addition to identifying strictly repeating motifs (gliding symmetry), we will explore strategies for extracting motifs for M.C. Escher's tessellation works by analyzing each of the seventeen symmetry groups \cite{schattschneider2010mathematical}. We also hope to deepen the interpretative capabilities of AI solutions by building the bridge between Gestalt principles of perception and algorithmic object manipulations for visual reasoning.  

\bibliographystyle{named}
\bibliography{ijcai20}

\begin{thebibliography}{}

\bibitem[\protect\citeauthoryear{Amalric \bgroup \em et al.\egroup
  }{2017}]{amalric2017language}
Marie Amalric, Liping Wang, Pierre Pica, Santiago Figueira, Mariano Sigman, and
  Stanislas Dehaene.
\newblock The language of geometry: Fast comprehension of geometrical
  primitives and rules in human adults and preschoolers.
\newblock {\em PLoS computational biology}, 13(1):e1005273, 2017.

\bibitem[\protect\citeauthoryear{Benesty \bgroup \em et al.\egroup
  }{2009}]{benesty2009pearson}
Jacob Benesty, Jingdong Chen, Yiteng Huang, and Israel Cohen.
\newblock Pearson correlation coefficient.
\newblock In {\em Noise reduction in speech processing}, pages 1--4. Springer,
  2009.

\bibitem[\protect\citeauthoryear{Bornstein \bgroup \em et al.\egroup
  }{1981}]{bornstein1981perception}
Marc~H Bornstein, Kay Ferdinandsen, and Charles~G Gross.
\newblock Perception of symmetry in infancy.
\newblock {\em Developmental psychology}, 17(1):82, 1981.

\bibitem[\protect\citeauthoryear{Carpenter \bgroup \em et al.\egroup
  }{1990}]{carpenter1990one}
Patricia~A Carpenter, Marcel~A Just, and Peter Shell.
\newblock What one intelligence test measures: a theoretical account of the
  processing in the raven progressive matrices test.
\newblock {\em Psychological review}, 97(3):404, 1990.

\bibitem[\protect\citeauthoryear{Dehaene \bgroup \em et al.\egroup
  }{2006}]{dehaene2006core}
Stanislas Dehaene, V{\'e}ronique Izard, Pierre Pica, and Elizabeth Spelke.
\newblock Core knowledge of geometry in an amazonian indigene group.
\newblock {\em Science}, 311(5759):381--384, 2006.

\bibitem[\protect\citeauthoryear{Escher and
  Schattschneider}{2004}]{escher2004visions}
Maurits~C Escher and Doris Schattschneider.
\newblock {\em Visions of Symmetry}.
\newblock Thames \& Hudson, 2004.

\bibitem[\protect\citeauthoryear{Forbus \bgroup \em et al.\egroup
  }{2011}]{forbus2011cogsketch}
Kenneth Forbus, Jeffrey Usher, Andrew Lovett, Kate Lockwood, and Jon Wetzel.
\newblock Cogsketch: Sketch understanding for cognitive science research and
  for education.
\newblock {\em Topics in Cognitive Science}, 3(4):648--666, 2011.

\bibitem[\protect\citeauthoryear{Haak}{1976}]{haak1976transformation}
Sheila Haak.
\newblock Transformation geometry and the artwork of {M}.{C}. {E}scher.
\newblock {\em Mathematics Teacher}, 69(8):647--652, 1976.

\bibitem[\protect\citeauthoryear{Koichu}{2008}]{koichu2008considerations}
Boris Koichu.
\newblock On considerations of parsimony in mathematical problem solving.
\newblock In {\em Proceedings of the 32nd Conference of the International Group
  for the Psychology of Mathematics Education}, volume~3, pages 273--280.
  Cinvestav-UMSNH Mexico, 2008.

\bibitem[\protect\citeauthoryear{Kunda \bgroup \em et al.\egroup
  }{2013}]{kunda2013computational}
Maithilee Kunda, Keith McGreggor, and Ashok~K Goel.
\newblock A computational model for solving problems from the raven’s
  progressive matrices intelligence test using iconic visual representations.
\newblock {\em Cognitive Systems Research}, 22:47--66, 2013.

\bibitem[\protect\citeauthoryear{Li}{2009}]{li2009interest}
Qi~Li.
\newblock Interest points of general imbalance.
\newblock {\em IEEE Transactions on Image Processing}, 18(11):2536--2546, 2009.

\bibitem[\protect\citeauthoryear{Lovett and Forbus}{2011}]{lovett2011cultural}
Andrew Lovett and Kenneth Forbus.
\newblock Cultural commonalities and differences in spatial problem-solving: A
  computational analysis.
\newblock {\em Cognition}, 121(2):281--287, 2011.

\bibitem[\protect\citeauthoryear{Lovett \bgroup \em et al.\egroup
  }{2008}]{lovett2008modeling}
Andrew Lovett, Kate Lockwood, and Kenneth Forbus.
\newblock Modeling cross-cultural performance on the visual oddity task.
\newblock In {\em International Conference on Spatial Cognition}, pages
  378--393. Springer, 2008.

\bibitem[\protect\citeauthoryear{McGreggor and
  Goel}{2011}]{mcgreggor2011finding}
Keith McGreggor and Ashok Goel.
\newblock Finding the odd one out: a fractal analogical approach.
\newblock In {\em Proceedings of the 8th ACM conference on Creativity and
  cognition}, pages 289--298. ACM, 2011.

\bibitem[\protect\citeauthoryear{McGreggor and
  Goel}{2013}]{mcgreggor2013fractal}
Keith McGreggor and Ashok Goel.
\newblock Fractal representations and core geometry.
\newblock In {\em Proceedings of the Second Annual Conference on Advances in
  Cognitive Systems ACS}, volume~3, page~19, 2013.

\bibitem[\protect\citeauthoryear{P{\'o}lya}{1990}]{polya1990mathematics}
George P{\'o}lya.
\newblock {\em Mathematics and plausible reasoning: Induction and analogy in
  mathematics}, volume~1.
\newblock Princeton University Press, 1990.

\bibitem[\protect\citeauthoryear{Santoro \bgroup \em et al.\egroup
  }{2018}]{santoro2018measuring}
Adam Santoro, Felix Hill, David Barrett, Ari Morcos, and Timothy Lillicrap.
\newblock Measuring abstract reasoning in neural networks.
\newblock In {\em International Conference on Machine Learning}, pages
  4477--4486, 2018.

\bibitem[\protect\citeauthoryear{Schattschneider}{2010}]{schattschneider2010mathematical}
Doris Schattschneider.
\newblock The mathematical side of m.c. escher.
\newblock {\em Notices of the AMS}, 57(6):706--718, 2010.

\bibitem[\protect\citeauthoryear{Shegheva and
  Goel}{2018}]{shegheva2018structural}
Snejana Shegheva and Ashok Goel.
\newblock The structural affinity method for solving the raven's progressive
  matrices test for intelligence.
\newblock In {\em Thirty-Second AAAI Conference on Artificial Intelligence},
  2018.

\bibitem[\protect\citeauthoryear{Spelke}{1990}]{spelke1990principles}
Elizabeth~S Spelke.
\newblock Principles of object perception.
\newblock {\em Cognitive science}, 14(1):29--56, 1990.

\bibitem[\protect\citeauthoryear{Tipping and
  Bishop}{1999}]{tipping1999probabilistic}
Michael~E Tipping and Christopher~M Bishop.
\newblock Probabilistic principal component analysis.
\newblock {\em Journal of the Royal Statistical Society: Series B (Statistical
  Methodology)}, 61(3):611--622, 1999.

\bibitem[\protect\citeauthoryear{Tyler}{1995}]{tyler1995empirical}
Christopher~W Tyler.
\newblock Empirical aspects of symmetry perception.
\newblock {\em Spatial Vision}, 9(1):1--8, 1995.

\bibitem[\protect\citeauthoryear{Wagemans \bgroup \em et al.\egroup
  }{2012}]{wagemans2012century}
Johan Wagemans, James~H Elder, Michael Kubovy, Stephen~E Palmer, Mary~A
  Peterson, Manish Singh, and R{\"u}diger von~der Heydt.
\newblock A century of gestalt psychology in visual perception: I. perceptual
  grouping and figure--ground organization.
\newblock {\em Psychological bulletin}, 138(6):1172, 2012.

\bibitem[\protect\citeauthoryear{Zadeh}{1984}]{zadeh1984making}
Lotfi~A Zadeh.
\newblock Making computers think like people [fuzzy set theory].
\newblock {\em IEEE spectrum}, 21(8):26--32, 1984.

\end{thebibliography}

\end{document}